%% file: main.tex
\documentclass[preprint,12pt,authoryear]{elsarticle}

\usepackage{amssymb}
\usepackage{amsmath}

\usepackage[noadjust]{cite}
\usepackage{natbib}
\usepackage{author_comments}
\usepackage{amsfonts}
\usepackage{algorithmic}
\usepackage{graphicx}
\usepackage{textcomp}
\def\BibTeX{{\rm B\kern-.05em{\sc i\kern-.025em b}\kern-.08em
    T\kern-.1667em\lower.7ex\hbox{E}\kern-.125emX}}
\usepackage{afterpage}
\usepackage{tabularx}
\usepackage{booktabs}
\usepackage{subcaption} 
\usepackage{hyperref}
\usepackage{hyphenat}
\usepackage{comment}

\usepackage[authoryear]{natbib}

\journal{Journal of Human-Computer Studies}

\begin{document}

\begin{frontmatter}


\author{Elmira Yadollahi\corref{cor1}\fnref{label1}}
\ead{elmiray@kth.se}
\cortext[cor1]{Authors contributed equally.}

\author{Fethiye Irmak Doğan\corref{cor1}\fnref{label1}}
\ead{fidogan@kth.se}

\author{Yujing Zhang\fnref{label1}}
\ead{yujingzh@kth.se}

\author{Beatriz Nogueira\fnref{label2}}
\ead{beatriz.c.nogueira@outlook.pt}

\author{Tiago Guerreiro\fnref{label2}}
\ead{tjvg@di.fc.ul.pt}

\author{Shelly Levy Tzedek\fnref{label3}}
\ead{shelly@bgu.ac.il}

\author{Iolanda Leite\fnref{label1}}
\ead{iolanada@kth.se}

\title{Expectations, Explanations, and Embodiment: Attempts at Robot Failure Recovery} 



\affiliation[label1]{organization={KTH Royal Institute of Technology},
             city={Stockholm},
             country={Sweden}}

\affiliation[label2]{organization={LASIGE, Faculdade de Ciências},
             city={Lisbon},
             country={Portugal}}

\affiliation[label3]{organization={Ben Gurion University of the Negev},
             city={Beer Sheva},
             country={Israel}}




\input{sections/00_Abstract}
\end{frontmatter}


\input{sections/00_Abstract.tex}
\input{sections/01_Introduction_new}
\input{sections/02_Related_Work_new.tex}
\input{sections/04_Validation_Priming_Videos.tex}

\input{sections/05_Main_Evaluation.tex}

\input{sections/06_Results.tex}
\input{sections/07_Discussion.tex}
\input{sections/09_Acknowledgment.tex}









\bibliographystyle{elsarticle-harv}
\bibliography{references}






\end{document}

%% file: sections/00_Abstract.tex
\begin{abstract}

Expectations critically shape how people form judgments about robots, influencing whether they view failures as minor technical glitches or deal-breaking flaws. This work explores how high and low expectations, induced through brief video priming, affect user perceptions of robot failures and the utility of explanations in HRI. We conducted two online studies ($N=600$ total participants); each replicated two robots with different embodiments, Furhat and Pepper. In our first study, grounded in expectation theory, participants were divided into two groups, one primed with positive and the other with negative expectations regarding the robot's performance, establishing distinct expectation frameworks. This validation study aimed to verify whether the videos could reliably establish low and high-expectation profiles. In the second study, participants were primed using the validated videos and then viewed a new scenario in which the robot failed at a task. Half viewed a version where the robot explained its failure, while the other half received no explanation. We found that explanations significantly improved user perceptions of Furhat, especially when participants were primed to have lower expectations. Explanations boosted satisfaction and enhanced the robot's perceived expressiveness, indicating that effectively communicating the cause of errors can help repair user trust. By contrast, Pepper's explanations produced minimal impact on user attitudes, suggesting that a robot's embodiment and style of interaction could determine whether explanations can successfully offset negative impressions. Together, these findings underscore the need to consider users' expectations when tailoring explanation strategies in HRI. When expectations are initially low, a cogent explanation can make the difference between dismissing a failure and appreciating the robot's transparency and effort to communicate.

\end{abstract}




%% file: sections/01_Introduction_new.tex
\section{Introduction}

When robots operate in human environments, user expectations play a crucial role in shaping human-robot interaction (HRI)~\citep{relatedwork_expectation_role_in_HRI, relatedwork_expectation_robot_interaction_skill, dogan2024grace}. However, there is often a mismatch between these expectations and the actual capabilities of social robots~\citep{relatedwork_expectation_evaluation_framework}, which can lead to disappointment and, consequently, diminish the quality of interactions~\citep{relatedwork_expectation_Expectancies,kruglanski2007principles}. For instance, a user might expect robots to function as proactive and autonomous assistants, yet when robots make mistakes due to their limited abilities, this mismatch can undermine the robot's perceived trustworthiness and competence~\citep{10.1145/2696454.2696497,cha2015perceived}. A promising approach for bridging this gap, i.e., aligning users’ expectations with the robot’s actual capabilities, can be through providing explanations for robot mistakes, which can improve users' trust towards robots~\citep{siau2018building, edmonds2019tale, barredoarrieta2020ExplainableArtificialIntelligence, doi:10.1177/15553434221136358} as well as the effectiveness of HRI~\citep{sridharan2019towards, setchi2020explainable}.

In social psychology, the mismatch between user expectations and actual system performance can be understood through the lenses of \textit{Attribution Theory}~\citep{weiner2010development} and \textit{Expectancy-Disconfirmation Theory}~\citep{oliver1994outcome}. 
According to Attribution Theory, individuals seek to interpret the causes of outcomes–especially failures–by assigning responsibility to internal or external factors. 
In the context of HRI, when a robot errs, users may attribute blame to the robot's inherent limitations or perceived incompetence. 
Meanwhile, Expectancy-Disconfirmation Theory posits that satisfaction hinges on whether actual performance meets or diverges from initial expectations. 
When there is a high expectation of a robot, its failure yields negative disconfirmation, diminishing trust and satisfaction of the robot. 
Both of these theoretical frameworks support the idea that providing explanations to clarify the robot's reasoning or constraints can help mitigate the negative effects of disappointment caused by the robot's failure or underperformance.



Previous research has highlighted the effects of user expectations on how people perceive robots \citep{relatedwork_expectation_role_in_HRI, relatedwork_expectation_robot_interaction_skill, relatedwork_expectation_evaluation_framework}, yet the impact of integrating robot explanations to handle the potential mismatch between user expectations and robot capabilities remains underexplored. Meanwhile, explanations, with their promising potential to recover from robot failures~\citep{das2021ExplainableAIRobot}, were delivered through visual~\citep{dougan2023leveraging, sobrin2024enhancing}, 
or verbal/textual forms~\citep{han2021need, stange2022self}, as well as incorporated into follow-up questions~\citep{9889368}. These studies have provided valuable insights into the impact of robot explanations while handling robot mistakes, but they have not considered individuals' expectations and preconceptions regarding robot capabilities during the explanation generation process. \textit{To address these open challenges, our study is the first to examine how people's expectations affect their perception of robot explanations.}

\begin{figure}[t]
     \centering
     \includegraphics[width=\linewidth]{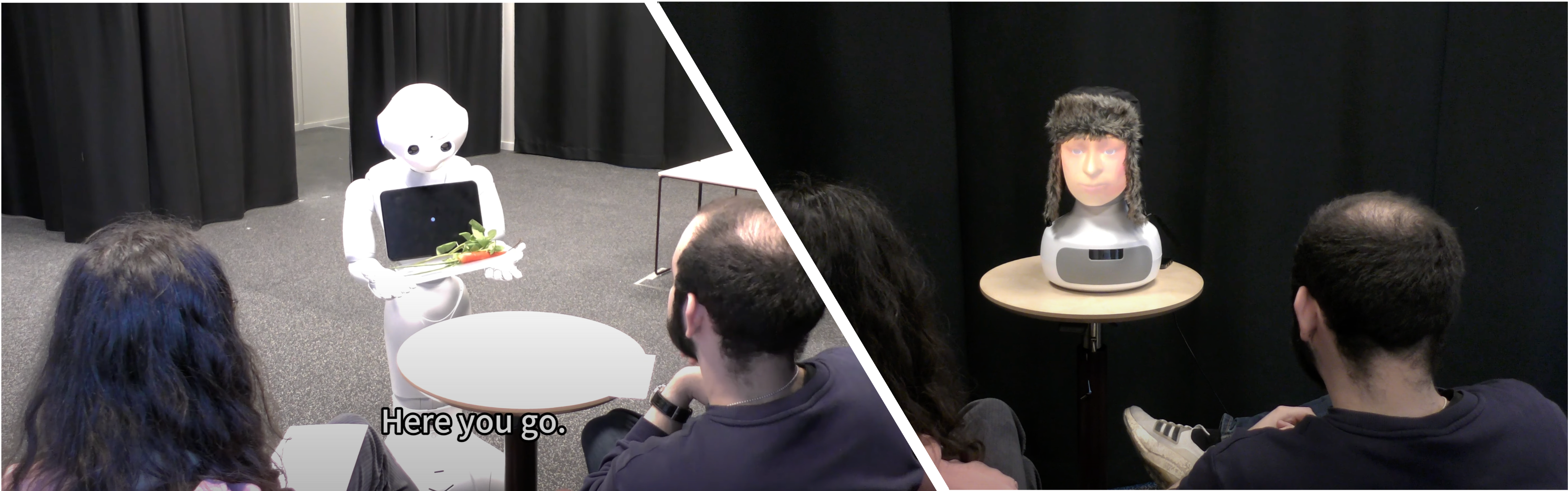}
     \caption{Priming scenario with Pepper (left) and Furhat (right).}
     \label{fig:video_setting}
\end{figure}

Our approach involves two distinct user studies, both replicated with two robots (Pepper and Furhat). In the first study (the ``priming study''), we aimed to validate whether short priming videos could reliably prime participants to hold either high or low expectations regarding robot capabilities (e.g., performing flawlessly versus making errors). Figure \ref{fig:video_setting} illustrates example scenes from these videos. The results showed that our priming method successfully induced the intended expectations. In the second study (the ``main evaluation''), we investigated how these primed expectations shaped user perceptions of robot explanations. Specifically, we exposed participants to scenarios where the robots deliberately made errors during a new task and either provided explanations for their mistakes or omitted any explanations. Our findings indicate that, across both low and high-expectation conditions, explanations generally improved the perception of the robots–particularly for Furhat. Notably, when participants held lower expectations, explanations had an even stronger positive effect, enhancing both the robot's perceived expressiveness and users' explanation satisfaction. 


%% file: sections/02_Related_work_new.tex
\nopagebreak
\section{Related Work}
\subsection{Role of Expectations}

``Expectation'' refers to the psychological concept that guides people's behaviour, hopes, and intentions \citep{relatedwork_expectation_Expectancies,kruglanski2007principles}.  A key framework for understanding how expectations shape perceptions and experiences is the Expectancy-Disconfirmation Theory (EDT), which suggests that individuals assess their satisfaction based on whether their expectations are met, exceeded, or unmet \citep{oliver1994outcome}. This process is particularly relevant in social contexts, including human-robot interaction, where individuals naturally form expectations that simplify the processing of familiar social situations \citep{relatedwork_expectation_situation_role_in_HRI} while making unexpected behaviours more challenging to interpret \citep{relatedwork_expectation_role_in_HRI}. 

In the context of HRI, previous research has highlighted that users expect robots to recognize and align with their expectations in various interaction roles \citep{relatedwork_expectation_situation_role_in_HRI}. Several studies have explored these expectations, focusing on robot appearance \citep{relatedwork_expectation_robot_appearance} and interaction abilities \citep{relatedwork_expectation_robot_interaction_skill}. Previous work has presented such expectations as dynamic concepts that can be changed based on several factors. For example, perceived interaction skills have been shown to shift depending on the user’s anticipated future role for the robot \citep{relatedwork_expectation_robot_interaction_skill}.


To examine how expectations impact interaction, Rosen et al. \citep{relatedwork_expectation_evaluation_framework} developed a framework for studying users' expectations of robots, focusing on affect, cognitive processing, and performance. However, there is still a mismatch between the expected and actual capabilities of robots, leading to potential disappointment and negative effects \citep{relatedwork_expectation_Expectancies,kruglanski2007principles,relatedwork_expectation_evaluation_framework}. Therefore, reducing the gap between expectations and reality is key to fostering long-term relationships, affecting users' evaluation with robots \citep{relatedwork_expectation_jokinen2017expectations}. 

\subsection{Role of Priming}

To overcome the expectation gap, priming offers a promising strategy by providing a significant impact on user expectations and behaviour \citep{relatedwork_priming_hri}. Priming is a non-conscious process associated with learning, where exposure to a priming stimulus influences the response to a subsequent target stimulus \citep{relatedwork_priming_hri}. For instance, ``movement priming'' refers to how one's movement can affect another person's actions or their own future movements \citep{relatedwork_priming_defination}.

Previous research has examined the influence of people's priming on HRI, highlighting how different forms of priming shape user perceptions, attitudes, and engagement with robotic systems. For example, \citet{relatedwork_priming_affectivepriming} explored how emotional priming impacts user perception in autonomous products, showing the potential to build long-term relationships via affective priming.
Additionally, research on media representations of robot characters has shown that sympathetic portrayals in media can prime positive social evaluations of robots, influencing individuals' mental models and social assessments \citep{relatedwork_priming_media}. 

\subsection{Role of Robot Failures}


As users' expectations are shaped either through natural interaction or priming, robot failures to meet their anticipations 
can significantly impact user satisfaction \citep{relatedwork_priming_hri}. 
In this context, Attribution Theory~\citep{weiner2010development} can provide useful insights for understanding how users interpret and react to these failures, as they may attribute them to either internal or external factors.


Following similar attributes, \citet{honig2018understanding} has identified two main categories of robot failures in HRI: technical and interaction failures, which are both implemented in our priming study. Technical failures typically stem from hardware malfunctions or issues in the robot's software system, like communication issues. On the other hand, interaction failures arise from uncertainties in interacting with humans or the environment, such as communication breakdowns and violations of social norms. 

Previous research has explored the impact of various robot failures in HRI \citep{honig2018understanding}, examining how different factors influence user responses and perceptions. For instance, \citet{relatedwork_failure_3374782} investigated user perceptions of conversational failures in robots, demonstrating that humanoid robots enhance users' responses to such failures. Moreover, other studies show that explanations enhance recovery from plan execution failures; for instance, Das et al. \citep{das2021ExplainableAIRobot} demonstrated that explanations incorporating context and prior actions are most useful for non-expert users in diagnosing failures and identifying solutions.

\subsection{Role of Explanations}

While handling robot failures, explanations have been shown to shape cognitive perceptions significantly and contribute to repairing users' mental models by potentially influencing users’ expectations via interactions \citep{relatedwork_explanation_MILLER20191, relatedwork_explanation_hilton1996mental}. A  well-designed explanation enhances transparency in the robot’s operation, improving user understanding, particularly for non-expert users \citep{relatedwork_explanation_hayes2017improving}. For instance, robots with explanations are often perceived as more lively and human-like \citep{relatedwork_explanation_9900558}, and they have been crucial for humans to understand robotic behaviour better \citep{relatedwork_explanation_han2021need}. On the other hand, an inadequate or unclear explanation can negatively affect user interaction \citep{relatedwork_explanation_3565480}.

Depending on the content being explained, robot explanations have been categorized into \textit{what-explanations}, \textit{why-explanations}, and \textit{how-explanations} \citep{relatedwork_explanation_MILLER20191}. Previous work has shown that clear \textit{why-explanations} are frequently required when robots behave unexpectedly, often perceived as failures \citep{relatedwork_explanation_3634990}, and such explanations are also leveraged during our study.

Despite existing research, there is a significant gap in understanding how robot explanations are perceived based on different user expectations and how such expectations are shaped by priming within HRI contexts. Therefore, further investigation is needed to assess the effect of priming on user expectations and to explore how robot explanations impact people's perceptions throughout the interaction. 

%% file: sections/04_Validation_Priming_Videos.tex
\nopagebreak
\section{Priming Study: Validating Priming Effect}
Before addressing our main research questions, we first sought to validate whether participants' perceptions of the robots could be influenced by short priming videos depicting robot interactions. 
Given that embodiment plays a crucial role in how people perceive and engage with robots, we decided to test our priming paradigm using two robots with distinctly different designs and interaction capabilities. 
We surveyed several commercially available robots used in research and narrowed our focus to \textit{Pepper} and \textit{Furhat} because they offer unique and contrasting embodiments and interaction experiences. 
\textit{Pepper} \citep{softbankpepper} is a humanoid robot equipped with a mobile base enabling it to move around and interact using verbal communication and body gestures. 
Despite this mobility and physical presence, Pepper's static face limits its expressiveness—its mouth doesn't move when speaking, and it cannot display facial emotions.
On the other hand, \textit{Furhat} \citep{furhatrobotics} is a stationary robot with a back-projected, human-like face capable of a wide range of nuanced facial expressions (e.g., eyebrow movements and nodding), which make it highly expressive. 
Although Furhat lacks mobility or adopting physical postures, its facial expressiveness allows for more intimate and personalized interactions. 

By selecting two robots that differ in embodiment, range of motion, and expressive modalities,  we aimed to explore whether these different embodiment profiles would influence (1) how easily participants could be primed (with positive and negative expectations) and (2) how participants might subsequently perceive or evaluate each robot. 
Prior research suggests that physical form heavily influences user engagement and perception. 
For example, Kiesler et al. \citep{kiesler2008anthropomorphic}  found that participants were more likely to anthropomorphize and engage with humanoid robots, suggesting that depending on what feature appeals to the user, different human-like features of each robot may make their explanations more effective.
Similarly, Li et al. \citep{li2017touching} showed that physical interaction with robots like Pepper can evoke emotional responses akin to those experienced during human interactions. 
These findings highlight that embodiment could shape how participants form expectations and interpret priming effects and explanations from each robot. Hence, we decided to replicate the study for both robots to identify the role of these inherent differences in robots in forming expectations and evaluating explanations.

\subsection{Study Objective}
Our primary goal was to design brief interaction scenarios with both Pepper and Furhat that would shift participants' perceptions of each robot's capabilities in either a positive or negative direction. 
We anticipated that after watching a corresponding priming video (positive or negative), participants' perceptions would differ significantly. Specifically:
\begin{itemize}[leftmargin=*]
    \item\textbf{Positive priming} videos highlighted \textbf{flawless} task execution and social interaction, timing to bolster confidence in the robot's competence.
    \item\textbf{Negative priming} videos showcased \textbf{failures} (communication failures, hardware malfunction, social norm violations) that would reduce confidence in the robot's competence. 
\end{itemize}
To evaluate the priming effect, we used scales measuring shared perception, interpretation, and nonverbal expressiveness (refer to section \ref{Measures}).
Prior research on priming in HRI supports the idea that priming can significantly alter users' perceptions of robots' abilities and behaviours \citep{eyssel2012s, song2023robot}. 
As Eyssel and Hegel \citep{eyssel2012s} found, positive priming improved participants' perceptions of a robot's likability and competence, supporting the idea that expectations can shape subsequent evaluations of robot behaviour and Song et al. \citep{song2023robot} demonstrated that emotional expressions and contextual cues enhanced perceptions of anthropomorphic trustworthiness in robots. 
These findings highlight that participants' preconceptions can be intentionally shaped to influence subsequent judgment of robot performance or trust.
In addition to these robot-focused measures, we also examined whether the priming videos could alter participants' general attitudes toward robots more broadly. 
As demonstrated by Mehrizi et al. \citep{mehrizi2022radiologists}, who found that attitudinal priming influenced radiologists' reliance on AI systems, there is an expectation that showing either positive or negative priming videos might shift participants' general perceptions of robotic technology.

\begin{table}[t!]
     \centering
     \caption{Video design for the priming study.}
     \label{tab:priming_video_design}
     \footnotesize
     \begin{tabularx}{0.98\textwidth}{|m{1.2cm} | m{1.3cm}|m{2.7cm} |m{2.3cm}|m{3.8cm} |}
         \toprule
        
         \textbf{Robot} & \textbf{Priming} & \textbf{\nohyphens{Communication failures}} & \textbf{\nohyphens{Hardware malfunctions}} & \textbf{\nohyphens{Social norm violations?}} \\ 
         \hline\hline
         \multirow{2}{=}{\textbf{Furhat}} 
         & \textit{Positive} 
         & \nohyphens{Speaks properly; correctly gets the answer}
         & \nohyphens{Normal prosody; completes the task}
         & \nohyphens{No - Keeps a proper social distance; response after user finished}\\
         \cline{2-5}
         & \textit{Negative} 
         & \nohyphens{Asks user to repeat the answer several times; fails to understand the verbal cue}
         & \nohyphens{Strange prosody; shuts down in the middle}
         & \nohyphens{Yes - Asks user to come closer and closer; Cuts user's response}\\ 
         \hline\hline
         \multirow{2}{=}{\textbf{Pepper}} 
         & \textit{Positive} 
         & \nohyphens{Speaks properly}
         & \nohyphens{Completes the task}
         & \nohyphens{No - Keeps a proper social distance; speaks in the correct direction}\\
         \cline{2-5}
         & \textit{Negative} 
         & \nohyphens{Fails to understand the verbal cue}
         & \nohyphens{Shuts down in the middle}
         & \nohyphens{Yes - Passes in the middle of two people talking; speaks with back towards the users}\\ 
         \bottomrule
     \end{tabularx}
     \vspace{-1em}
 \end{table}

\subsection{Design of Priming Videos}

The priming videos were designed to clearly depict each robot's capabilities or potential failures to shape participants' expectations of the robot. 
Both robots were placed in a restaurant setting and chosen to provide a relatable, everyday scenario: 

\begin{itemize}[leftmargin=*]
    \item[-] \textbf{Pepper} was portrayed as a waiter taking orders and interacting with customers.
    \item[-] \textbf{Furhat} was portrayed as a customer satisfaction agent, asking patrons about their experience.
\end{itemize}

Following Honig et al. \citep{honig2018understanding}, the \textbf{negative} videos included communication failures (e.g., not understanding verbal cues), hardware malfunctions (e.g., shutting down mid-task), and social norm violations (e.g., interrupting or standing too close). The \textbf{positive} videos used the same general scenarios but showed the robots executing tasks seamlessly and interacting appropriately. Each video lasted approximately two minutes, with subtitles to ensure clarity. Table \ref{tab:priming_video_design} details the specific failures and successes depicted. 

Drawing on the failure taxonomy by Honig et al. \citep{honig2018understanding}, we incorporated three failure types for the negative priming videos—\textit{communication failures}, \textit{hardware failures}, and \textit{social norm violations}—as these were directly attributable to the robot and were visually demonstrable in video format.
The failures were tailored to each robot’s unique capabilities, but the scenarios were kept as comparable as possible to ensure consistency across groups. 
The positive priming videos followed the same structure but without any failures, showcasing the robots performing tasks flawlessly.
Each video lasted approximately two minutes, and subtitles were included for clarity. The specific scenarios used in the videos are detailed in Table \ref{tab:priming_video_design}.





\subsection{Priming Study Design}
We developed a between-subjects study with two priming conditions (\textit{positive}, \textit{negative}) repeated over for two robot types (\textit{Furhat}, \textit{Pepper}) design. 
This resulted in the development of four priming videos (Furhat-positive $\times$ Furhat-negative $\times$ Pepper-positive $\times$ Pepper-negative), where participants were randomly assigned to.
After viewing the video, participants completed questionnaires measuring their perception of the robot (shared perception, interpretation, nonverbal expressiveness) and general attitudes toward robots. By comparing scores across conditions, we could infer the effectiveness of the positive vs. negative priming for each robot. 



\subsection{Participants and Power Analysis}
\label{Participant}
We recruited participants ($N=208$) through Prolific platform \citep{palan2018prolific}, following an \textit{a priori power analysis} using G*Power \citep{faul2009statistical} that indicated a need for 144 participants (72 per robot) to detect a large effect ($f = 0.5$) with $\alpha$ (error probability) $=0.1$ and Power ($1=\beta$ error probability) $=0.9$.
Based on an initial pilot with 20 participants, we estimated the study would take around 12 minutes, and participants were compensated with 1.8£ at an hourly rate of 9£ per hour. The median completion time for this study was 10 min 30 s. 
The Furhat priming videos lasted for 2 min 45 s (\textit{negative priming}) and 3 min (\textit{positive priming}), while the Pepper priming videos lasted 2 min 18 s (\textit{negative priming}) and 2 min 30 s (\textit{positive priming}). 
The participant pool was set to all available countries, with the following pre-screening metrics: fluent in English, approval rate of 98-100, and having 20-10,000 previous submissions. A total of 8 participants failed the attention check question and were excluded from the study. 

Out of 200 participants, 106 identified as female, 91 as male, two as non-binary, and one preferred not to say. The participants' age ranged from 18 to 67 ($Mdn = 29.49\pm8.79$). 
Finally, we also requested participants to select their level of interaction with robots on a scale from ``No experience'' to ``I work with robots daily''.  
From 200 responses, 75 selected ``I have interacted with a robot'', 68 mentioned ``I have seen a robot'', 34 had ``No experience'', 22 specified ``I had multiple interactions with robots'', and one said ``I work with robots daily''. 

\subsection{Measures}
\label{Measures}
\subsubsection{The Peculiarities of Robot Embodiment (EmCorp-Scale)}
The EmCorp scale, developed and validated by Hoffmann et al. \citep{Emcorp}, provides a theoretical framework assessing users’ perceptions of artificial entities’ bodily-related capabilities. 
We used a modified version of the 7-point Likert EmCorp-Scale, focusing on three constructs: (1) \textit{Shared Perception and Interpretation}, (2) \textit{Tactile Interaction and Mobility}, and (3) \textit{Nonverbal Expressiveness}. We excluded the \textit{Corporeality} construct, which represents the robot's co-presence with the observer–an aspect not under investigation in this study. All items were rated on a 7-point Likert scale from ``strongly disagree'' to ``strongly agree.'' The details of the constructs we used are as follows. 

\begin{itemize}[leftmargin=*]
   \item \textbf{(Shared) Perception and Interpretation} (9 items): Assesses the robot’s perceived perceptual capabilities, including vision and hearing. In the text, we refer to this construct as \textit{Interpretation}.
   \item \textbf{Tactile Interaction and Mobility} (8 items): Measures the robot’s perceived ability to move around, manipulate objects, and generally function in physical space. We refer to this as \textit{Mobility}.
   \item \textbf{(Nonverbal) Expressiveness} (4 items): Captures the robot’s ability to convey meaning through natural cues such as gestures and facial expressions. We refer to this as \textit{Expressiveness}.
\end{itemize}

\subsubsection{General Attitudes Towards Robots Scale (GAToRS)}
GAToRS is a multidimensional scale developed and validated by Koverola et al. \citep{GATORS} that measures people’s positive and negative attitudes, giving them equal weight. 
It comprises 20 items rated on a 7-point Likert scale (1 = ``\textit{strongly disagree}'', 7 =``\textit{strongly agree}''). These items are distributed across four subscales, each focusing on personal- and societal-level attitudes:

\begin{itemize}[leftmargin=*]
   \item \textbf{Personal level positive ($P+$)}  (5 items): Assesses trust, comfort, and overall feeling of ease towards robots, persons, and organizations related to their development. 
   \item \textbf{Personal level negative ($P-$)} (5 items): Captures feelings of unease, fear, and nervousness around robots.
   \item \textbf{Societal level positive ($S+$)} (5 items) Evaluates the perceived benefits of robots within broader societal contexts (e.g., work, society, daily life).
   \item \textbf{Societal level negative ($S-$)} (5 items) Assesses concerns about robots' societal impacts on people's lives, jobs, and society (e.g., job displacement, privacy issues).
\end{itemize}

\begin{figure}[t!]
     \centering
     \includegraphics[width=\linewidth]{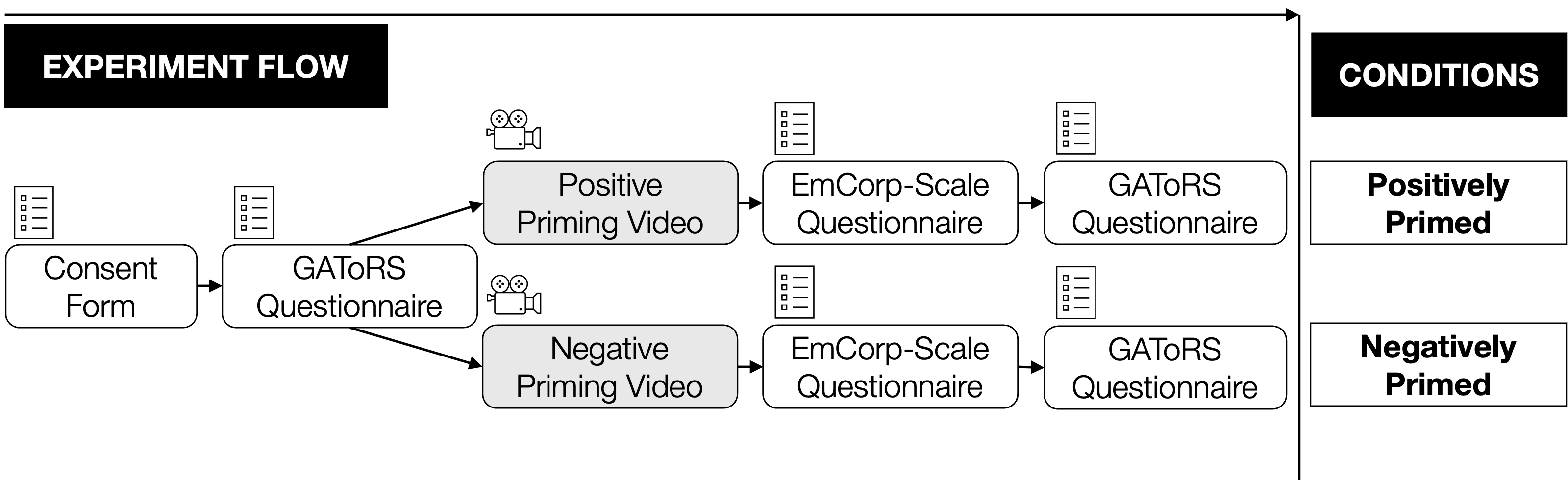}
     \caption{Experiment flow for the priming study.}
     \label{fig:priming_flow}
     \vspace{-1em}
\end{figure}

\subsection{Study Procedure}
Participants were recruited via Prolific and were redirected to a Qualtrics survey. 
After providing informed consent, they completed demographic questions with additional items covering their familiarity or interaction level with robots. 
Next, they answered pre-GAToRS~\citep{GATORS}, establishing their baseline attitudes toward robots. 
They were then randomly assigned to watch one of the priming videos (positive or negative; Pepper or Furhat), after which they rated their impression of the robot using the EmCorp scale~\citep{Emcorp}. 
Finally, participants repeated the GAToRS questionnaire (post-GAToRS), allowing us to gauge any change in their general attitudes following the priming video.
The overall study flow is shown in Figure~\ref{fig:priming_flow}.
At this stage, we incorporated a simple attention check question within the EmCorp questionnaire to examine whether they were legitimately paying attention to the questions.
An attention-check question was embedded in the EmCortp questionnaire to ensure data quality. The question expected participants to respond with a disagree when they agreed with a statement.

\subsection{Priming Results}

\subsubsection{Data Preparation}
We computed Cronbach's alpha for each subscale to assess the internal consistency, considering values above 0.7 acceptable and above 0.6 marginally acceptable. 
For EmCorp, we had the following Cronbach's alpha for \textit{interpretation} ($\alpha \geq 0.8$), \textit{mobility} ($\alpha \geq 0.8$), and \textit{expressiveness} ($\alpha \geq 0.6$).
For GAToRS, we had the following Cronbach's alpha for the pre-GAToRS and post-GAToRS collection, respectively: ($\alpha_{pre} \& \alpha_{post} \geq 0.6)$ for \textit{personal positive}, \textit{societal positive}, and \textit{societal negative} subscale, and ($\alpha_{pre} \& \alpha_{post} \geq 0.7)$ for \textit{personal negative} subscale. 
In the EmCorp scale, although \textit{mobility} met acceptable alpha levels, we excluded it from further analysis because Furhat's lack of mobility made the use of this subscale unsuitable for Furhat, and hence it was excluded for both robots. 

\subsubsection{Effect of Priming on Robot Perception}
\begin{figure*}[t!]
     \centering
     \begin{minipage}[b]{0.4\textwidth}
         \centering
         \includegraphics[width=\linewidth]{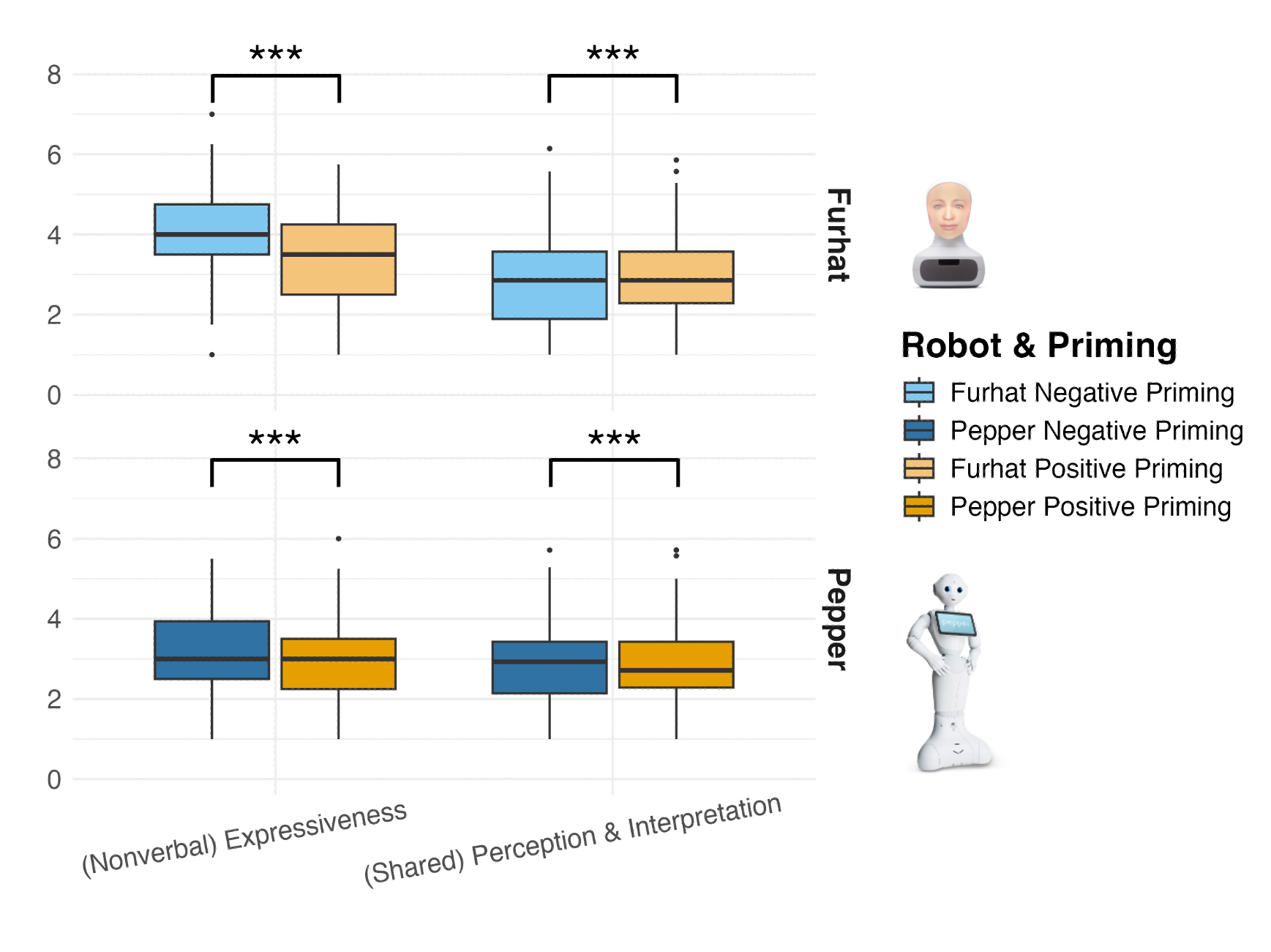}
         \caption{EmCorp results for the priming study.}
         \label{fig:Emcorp_priming}
     \end{minipage}
     \hfill
     \begin{minipage}[b]{0.59\textwidth}
         \centering
         \includegraphics[width=\linewidth]{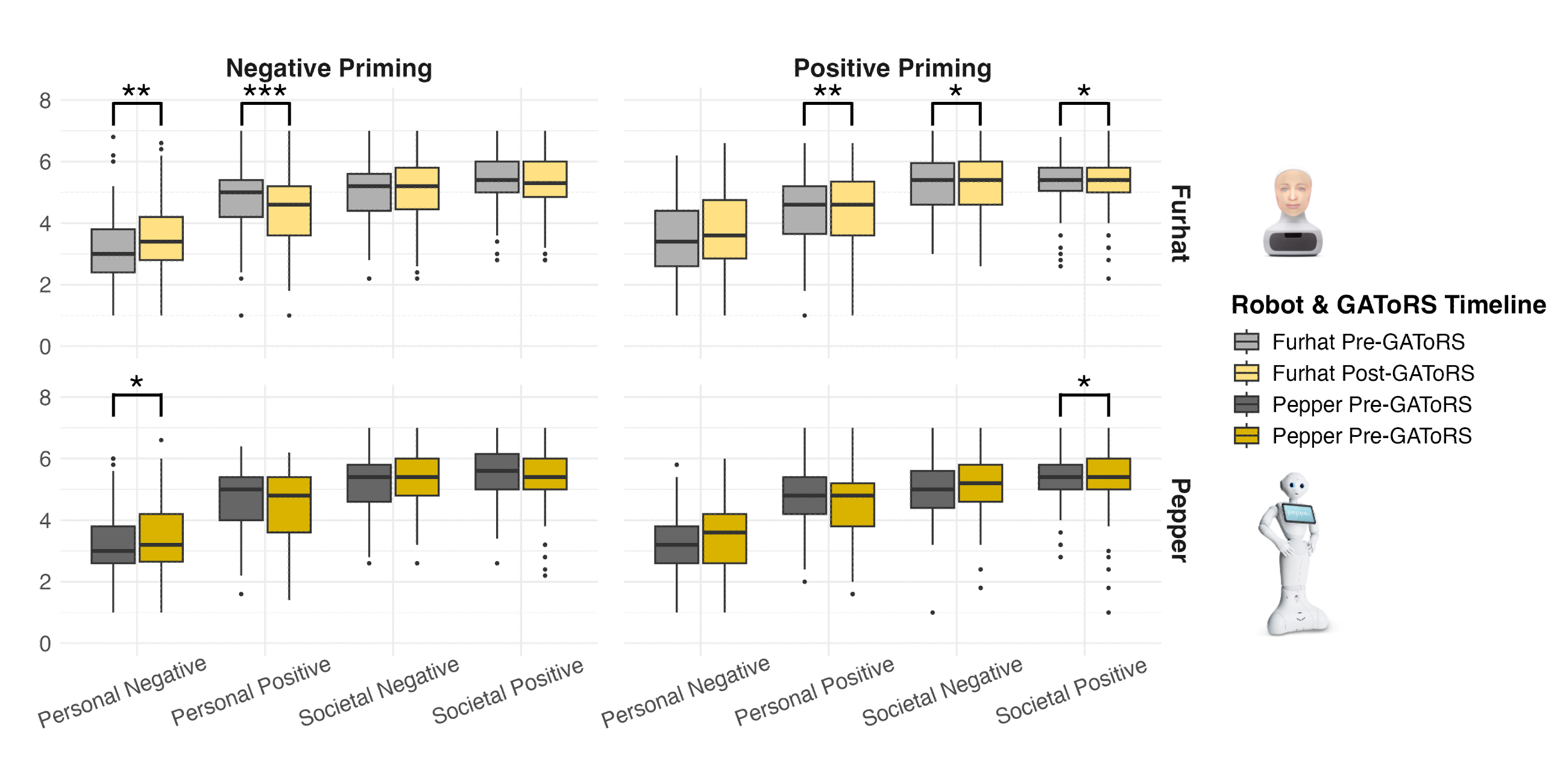}
         \caption{GAToRS results for the priming study.}
         \label{fig:Gators_priming}
     \end{minipage}
     \vspace{-2em}
 \end{figure*}

We conducted Wilcoxon signed-rank tests on the EmCorp subscales to examine whether the positive versus negative priming videos led to distinct perceptions of the same robot in similar scenarios.
Figure~\ref{fig:Emcorp_priming} illustrates the median scores. 


\textbf{Furhat:}
The Wilcoxon signed-rank test indicated that the robot's \textit{interpretation} was rated significantly higher $W = 507.5, p = 5.244e-07$ in the positive priming condition ($Mdn = 4.0\pm1.08$) compared to negative priming ($Mdn = 2.85\pm0.99$).
The robot's \textit{expressiveness} was also rated significantly higher $W = 749.5, p = 0.000867$ when participants were positively primed ($Mdn = 4.0\pm0.91$) compared to negative priming condition ($Mdn = 3.75\pm0.90$).

\textbf{Pepper:}
The Wilcoxon signed-rank test indicated that Pepper's \textit{perceptions and interpretation} abilities were rated significantly higher $W = 655.5, p = 2.558e-05$ when participants were positively primed ($Mdn = 3.71\pm1.09$) and compared to negative priming condition ($Mdn = 2.71\pm1.09$).
In terms of rating of Pepper's \textit{expressiveness}, the Wilcoxon signed-rank test shows that when participants were positively primed ($Mdn = 4.0\pm0.95$), they rated it significantly higher $W = 424, p = 6.969e-09$ compared to when they were negatively primed ($Mdn = 2.6\pm1.02$). 

\subsubsection{Effect of Priming on General Attitudes}
We next examined how priming might influence participants' broader general attitudes toward robots by looking at changes from pre-GAToRS to post-GAToRS. Figure~\ref{fig:Gators_priming} illustrates these subscale changes. 

\textbf{Furhat:}
When participants were \textbf{negatively primed}, we only observed significant changes in Personal Negative and Personal Positive subscales. 
\begin{itemize}[leftmargin=*]
    \item $\uparrow$ in Personal Negative: $W = 254, p = 0.002$; pre ($Mdn = 3.2\pm1.02$) $\rightarrow$ post ($Mdn = 3.6\pm1.22$)
    \item $\downarrow$ in Personal Positive: $W = 825, p = 0.00011$; pre ($Mdn = 4.8\pm0.93$) $\rightarrow$ post ($Mdn = 4.5\pm1.06$)
\end{itemize}

When participants were \textbf{positively primed}, we only observed significant changes in Personal Negative, Societal Positive, and Societal Negative subscales. 

\begin{itemize}[leftmargin=*]
    \item $\uparrow$ in Personal Negative: $W = 132, p = 0.002$; pre ($Mdn = 4.6\pm0.98$) $\rightarrow$ post ($Mdn = 4.8\pm1.00$)
    \item $\uparrow$ in Societal Positive: $W = 132, p = 0.002$; pre ($Mdn = 5.4\pm0.79$) $\rightarrow$ post ($Mdn = 5.4\pm0.84$)
    \item $\uparrow$ in Societal Negative: $W = 200, p = 0.012$; pre ($Mdn = 5.0\pm0.95$) $\rightarrow$ post ($Mdn = 5.2\pm0.97$)
\end{itemize}

\textbf{Pepper:}
With respect to the Pepper robot, when participants were \textit{negatively primed}, we only observed a significant change in one of the subscales, Personal Negative. 
\begin{itemize}[leftmargin=*]
    \item $\uparrow$ in Personal Negative: $W = 195, p = 0.018$; pre ($Mdn = 3.0\pm1.15$) $\rightarrow$ post ($Mdn = 3.2\pm1.03$)
\end{itemize}

On the other hand, when they were \textit{positively primed}, a significant change was only observed in the Societal Positive subscale. 
\begin{itemize}[leftmargin=*]
    \item $\uparrow$ in Societal Positive: $W = 264.5, p = 0.031$; pre ($Mdn = 5.2\pm0.84$) $\rightarrow$ post ($Mdn = 5.6\pm0.84$)
\end{itemize}



\subsection{Validating Priming Effect: Discussion}
Our findings demonstrate that priming can meaningfully shift how participants perceive robot capabilities in terms of \textit{interpretation} and \textit{expressiveness}. 
When participants viewed positive priming videos, both Furhat and Pepper were rated higher for these characteristics, consistent with the idea that setting higher expectations can enhance perceived competence–even when the robot's behaviour remains the same.

Furthermore, general attitudes toward robots also changed following priming.
For Furhat, negative priming increased \textit{personal negative} scores while decreasing \textit{personal positive} attitudes; conversely, positive priming raised both \textit{personal positive} and \textit{societal positive} attitudes. Pepper's negative priming significantly elevated personal negative scores, whereas positive priming improved \textit{societal positive} attitudes. Thus, priming not only affects how users appraise specific robot capabilities but also reshapes broader, more stable attitudes toward robots–an outcome that has important implications for expectation management and designing interactions to mitigate the impact of robot shortcomings. 
Still, it is important to note that participants completed these questionnaires immediately after viewing the priming videos, so the longevity of these priming effects remains uncertain. Future research could investigate whether repeated or prolonged interactions might sustain (or erode) these altered expectations over time. 





%% file: sections/05_Main_Evaluation.tex
\section{Main Evaluation: Effect of Failures and Explanation}

In this study, we built upon our validated priming approach to explore how people's expectations–shaped by priming videos–impact their perception of robot failures and subsequent explanations. 
Specifically, we investigated the role of explanations in recalibrating participants' expectations and influencing their overall perception of the robots over a short-term interaction. We did not collect longitudinal data, so the persistence of these effects remains an open question. 

\subsection{Hypotheses}

\textbf{H1 (Perception $\times$ Explanation):} \textit{Participants' perceptions of the robot, measured by the EmCorp scale, will improve significantly after receiving an explanation of failure, regardless of whether they were negatively or positively primed.}

This hypothesis is supported by Eyssel et al. \citep{eyssel2012s}, who found that explanations play a pivotal role in shaping perceptions, especially when users initially misinterpret a robot’s functionality based on its form. 
Additionally, de Visser et al. \citep{de2020towards} demonstrated that explanations can enhance perceived competence and restore trust in robots, even following failures.




\textbf{H2 (Explanation Satisfaction):} \textit{Participants who were negatively primed will report higher satisfaction after explanations of robot failures.}

Prior research suggests that priming may influence users' baseline expectations, thereby affecting how they respond to robotic failures. For example, Haring et al. \citep{haring2014perception} found that cultural and situational priming can modulate user perceptions of trustworthiness and satisfaction with robots.
Meanwhile, Salem et al. \citep{salem2012generation} demonstrated that providing explanations for robot errors can enhance user satisfaction and perceived trust, especially when initial expectations of the robot's capabilities are low, as in the case of negative priming.
The study showed that when robots communicate their failures effectively, users are more forgiving and more likely to perceive the robot as competent and trustworthy, even after the occurrence of failure.

\subsection{Design of Failure Videos with and without Explanations}
The videos for the main evaluation were designed to mirror the style and structure of the priming videos while portraying new tasks and different failure instances. Additionally, these videos featured new actors and distinct scenario settings to maintain a clear differentiation from priming ones. The main videos took place in a museum:
\begin{itemize}[leftmargin=*]
    \item \textbf{Pepper} served as a museum guide
    \item \textbf{Furhat} acted as a robot involved in rating the service
\end{itemize}

Similar to the negative priming videos (as they showcased failure cases), each main task video included three types of errors–communication failures, hardware malfunctions, and social norm violations–with the distinct museum scenarios presented in Table \ref{tab:main_video_design}.
For Pepper, we introduced a visibly disruptive failure (dropping an object), reflecting its more extensive physical interaction capabilities. Furhat's errors remained centred on communication lapses, consistent with its stationary form factor. 

When an explanation was provided, the robot apologized with a brief, one-sentence statement explaining the cause of the failure. 
This approach lets us examine how explanations alone, absent of task success, might influence participants' willingness to forgive errors and modulate their perception of the robot. In videos without an explanation, the robot did not address its failure. 
 \begin{table}[t!]
     \centering
     \caption{Video design for the main study.}
     \label{tab:main_video_design}
     \footnotesize
     \begin{tabularx}{0.98\textwidth}{| m{1.3cm} | m{4.6cm} | m{6.2cm} |}
         \toprule
         \textbf{Robot} & \textbf{Failure} & \textbf{Explanation} \\ 
         \hline\hline
         \multirow{3}{=}{\textbf{Furhat}} 
         & Looks in the other direction when speaking
         & Sorry, I couldn't recognize where you are because you are too far away\\
         \cline{2-3}
         & Speaks over a person
         & Sorry, I couldn't recognize you were still talking because your voice is too low\\
         \cline{2-3}
         & Talks to a person and the voice gets un-understandable
         & Sorry, I am having problems with my motor functions, which is affecting my speech \\
         \hline\hline
         \multirow{3}{=}{\textbf{Pepper}} 
         & Looks in the other direction when speaking
         & Sorry, I couldn’t recognize where you are because there is a problem with my camera.\\
         \cline{2-3}
         & Speaks over a person
         &  Sorry, I couldn’t recognize you were still talking because your voice is too low.\\
         \cline{2-3}
         & Drops an object while carrying it
         & "Sorry, the object was too heavy for me to carry"\\
         \bottomrule
     \end{tabularx}
     \vspace{-1em}
\end{table}

\subsection{Study Design}
\begin{figure*}[t!]
     \centering
     \includegraphics[width=\linewidth]{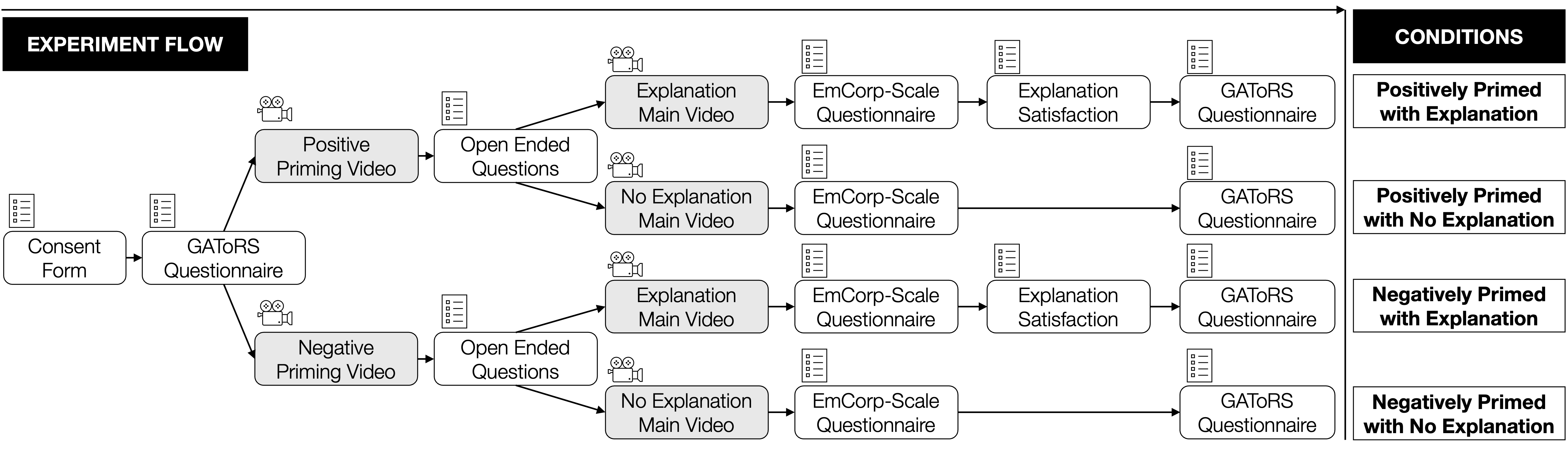}
     \caption{Experiment flow for the main study.}
     \label{fig:main_flow}
     \vspace{-1.4em}
 \end{figure*}

We developed a between-subjects study with a 2 (Priming Type: \textit{Positive}, \textit{Negative}) $\times$ 2 (Explanation Type: \textit{Explanation}, \textit{No Explanation}) design.  
Additionally, the study was replicated for two different robot embodiments (Robot Type: \textit{Furhat}, \textit{Pepper}), leading to a total of eight video stimuli. 
As illustrated in Figure \ref{fig:main_flow}, participants first watched one of the priming videos (in a restaurant setting) to establish high or low expectations. They then viewed a second video (in a museum setting) in which the same robot encountered failures, and the interaction was either accommodated with explanations after each failure or explanations were not provided by the robot. 

\subsection{Participants and Power Analysis}
To determine the sample size, we conducted an \textit{a priori} power analysis using G*Power \citep{faul2009statistical}: Goodness-of-fit test. 
We used the following parameters: Degree of Freedom $=3$, a large effect size $f=0.5$, $\alpha$ (error probability) $=0.1$, Power ($1-\beta$ error probability)$=0.9$. 
The selected parameter estimated that 50.48 participants per condition $\sim$ 200 participants are needed to achieve the expected results. 
For the two robots, we recruited $N=455$ participants through the Prolific platform \citep{palan2018prolific}.
Based on an initial pilot with 20 participants, we estimated the study would take around 18 minutes, and participants were compensated with 2.7£ at an hourly rate of 9£ per hour. The median completion time for this study was 18 min 34 s. 
Each participant watched two videos in this study; the duration of priming videos is reported in Section \ref{Participant}, and the lengths of the main videos are as follows: Furhat videos lasted 1 min 40 s (no explanation) and 2 min 10 s (explanation), and the Pepper videos lasted for 2 min (no explanation) and 2 min 18 (explanation). 
We used the same pre-screening metrics as the priming validation study. 

After removing 40 participants who failed attention checks, we had 415 valid responses (239 female, 172 male, and four non-binary). The participants' age ranged from 18 to 67 ($Mdn = 31.80\pm10.48$). 
Participants' level of interaction with robots is as follows: 142 selected ``I have interacted with a robot'', 136 mentioned ``I have seen a robot'', 75 had ``No experience'', 59 specified ``I had multiple interactions with robots'', and three said ``I work with robots daily''. 

\subsection{Measures}
We collected the same EmCorp (Peculiarities of Robot Embodiment) and GAToRS (General Attitudes Towards Robots Scale) data as in the priming study. Additionally, for participants in the explanation condition, we used an extra questionnaire to collect their explanation satisfaction.


\subsubsection{Explanation Satisfaction Questionnaire}
Explanation satisfaction was measured using a scale developed by Hoffman et al. (2018) \citep{Emcorp}. The scale comprises eight items, each rated on a fully labelled 5-point Likert scale (1 = ``strongly disagree'', 5 = ``strongly agree''). Originally designed to assess explanations provided by software, algorithms, or tools, the scale was adapted in this study to evaluate explanations from robots.

\subsection{Study Procedure}
The main study followed a similar structure to the priming evaluation, with the addition of another round of videos and questionnaires. Figure~\ref{fig:main_flow} outlines the entire procedure. 
Upon being redirected to the Qualtrics survey, participants consented and completed demographic/robot familiarity questions and were briefed about the flow of the study.
Participants completed pre-GAToRS \citep{GATORS} to establish their baseline attitudes and filled the post-GAToRS as the final step of the experiment. 
After pre-GAToRS, they watched one priming video (positive or negative), maintaining consistent conditions from the priming validation. 
Unlike in the priming-only evaluation, we did not administer EmCorp immediately after this first video. 
Instead, we inserted attention checks and open-ended queries about the scenario, giving participants a short break before proceeding. 
Participants next viewed the main task video (failures, with or without explanations). 
Those in the explanation condition encountered the robot's explanation for each failure; those in the no explanation condition saw the same failures without any clarifications. 
We then administered the EmCorp scale to assess how users perceived the robot following its mistakes. 
If participants received explanations, they also filled out the explanations satisfaction questionnaire. 
Finally, participants completed a post-GAToRS to capture any shifts in their general attitudes after priming their expectations and witnessing robot failures in the subsequent videos. 
Notably, both videos featured the same robot (Pepper or Furhat) to ensure continuity of embodiment. However, the tasks were deliberately different (restaurant vs. museum settings) to minimize boredom and clarify that the robot might fail or succeed in multiple contexts. As in the priming study, a semi-wizarded operation was used to time the failures and explanations appropriately. 



%% file: sections/06_Results.tex
\vspace{-1em}
\section{Main Evaluation Results}

\subsection{Data Preparation}
Similar to the priming study, we calculated Cronbach's alpha for all the questionnaires and excluded the \textit{mobility} subscale of EmCorp.
For EmCorp, Cronbach's alpha of \textit{interpretation} ($\alpha \geq 0.8$) and \textit{expressiveness} ($\alpha \geq 0.6$) both showed acceptable internal consistency. 
For GAToRS, we computed Cronbach's alpha for pre- and post collections: ($\alpha_{pre} \& \alpha_{post} \geq 0.7)$ for \textit{personal positive}, \textit{personal negative}, and \textit{societal positive} subscale, and ($\alpha_{pre} \& \alpha_{post} \geq 0.6)$ for \textit{societal negative} subscale. 
Finally, Cronbach's alpha of the \textit{explanation satisfaction} scales yielded to ($\alpha \geq 0.8$).

\subsection{Baseline and Post-Study General Attitudes (GAToRS)}

We used the GAToRS to gauge how participants felt about robots before (pre-GAToRS) and after (post-GAToRS) the main study. Our goal was to see whether participants in the four experimental conditions–(\textit{positive-explanation}, \textit{positive-no explanation}, \textit{negative-explanation}, \textit{negative-no explanation})–started with similar or differing attitudes, and whether those attitudes shifted by the end of the study. 

\textbf{Furhat:}
With respect to \textit{Baseline (Pre-GAToRS)} scores, given that the data was measured on Likert-type scales, we used the non-parametric Kruskal-Wallis test to compare pre-GAToRS scores across the four conditions. Results showed a statistically significant difference in the Personal Negative subscale ($H(3) = 10.78, p = 0.012$), indicating that at least one group entered the study feeling more negatively about robots compared to the others. However, no significant differences were observed for the Personal Positive, Societal Negative, or Societal Negative subscales ($p > 0.05$).

With respect to \textit{Post-Study (Post-GAToRS)} responses, which were collected after participants watched the priming and main videos featuring the robot failures and (in some conditions) explanations. The same Kruskal-Wallis test revealed no significant differences across the four conditions ($p > 0.1$) on any GAToRS subscale. This suggests that, by the end of the study, whatever gap existed at baseline (specifically in Personal Negative) had effectively levelled out, leaving participants with no statistically distinguishable differences in their general attitudes across conditions. One possible interpretation is that exposure to the main study videos–even with variations in failures and explanations–brought participants' negative or positive attitudes closer together, effectively normalizing their views about robots. 


\begin{figure*}[t!]
     \centering
     \begin{minipage}[b]{0.59\textwidth}
 \includegraphics[width=\linewidth]{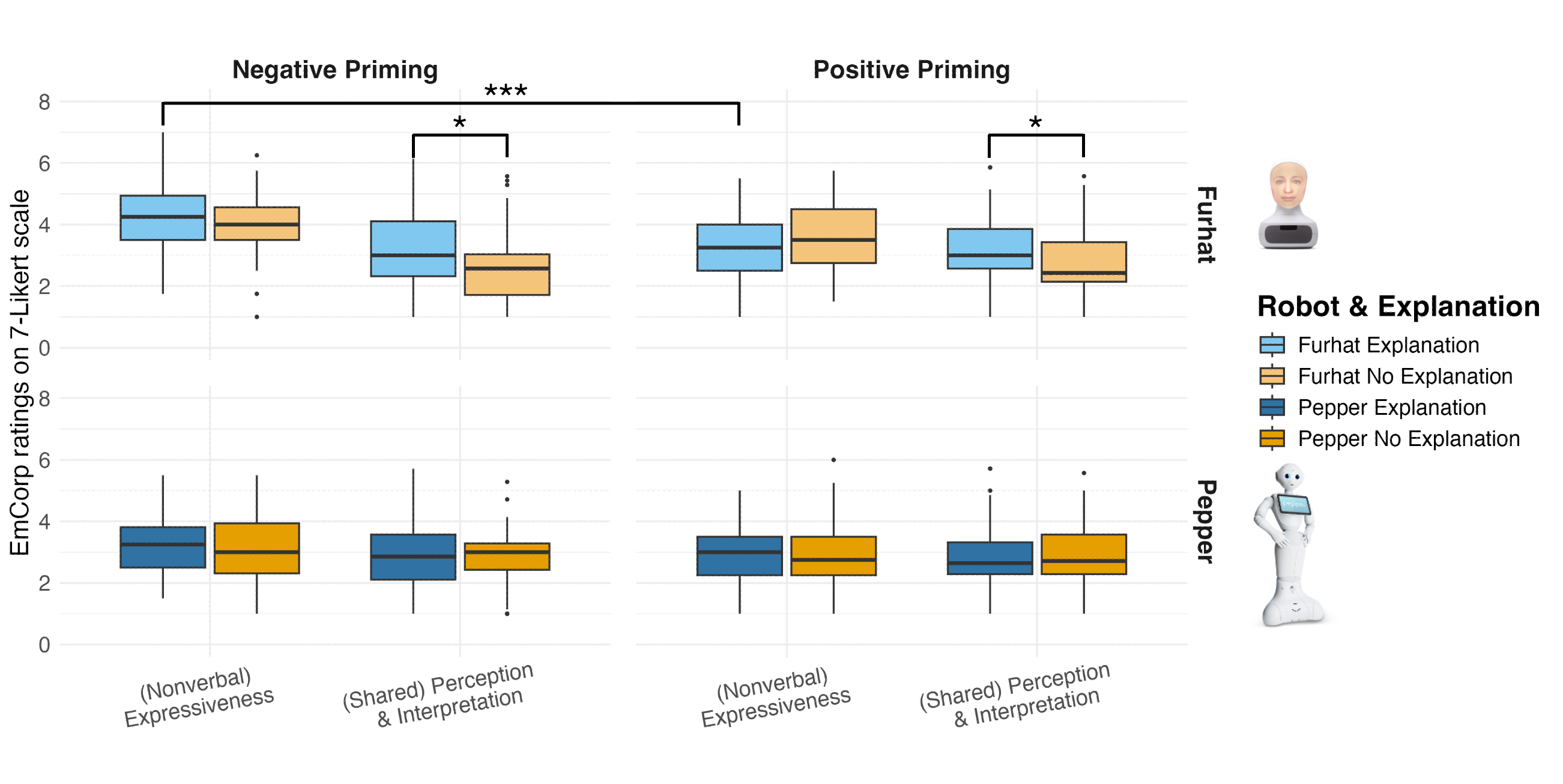}
         \caption{EmCorp results for \\the main study.}
         \label{fig:Emcorp_main}
     \end{minipage}
         \begin{minipage}[b]{0.39\textwidth}
         \centering
         \includegraphics[width=\linewidth]{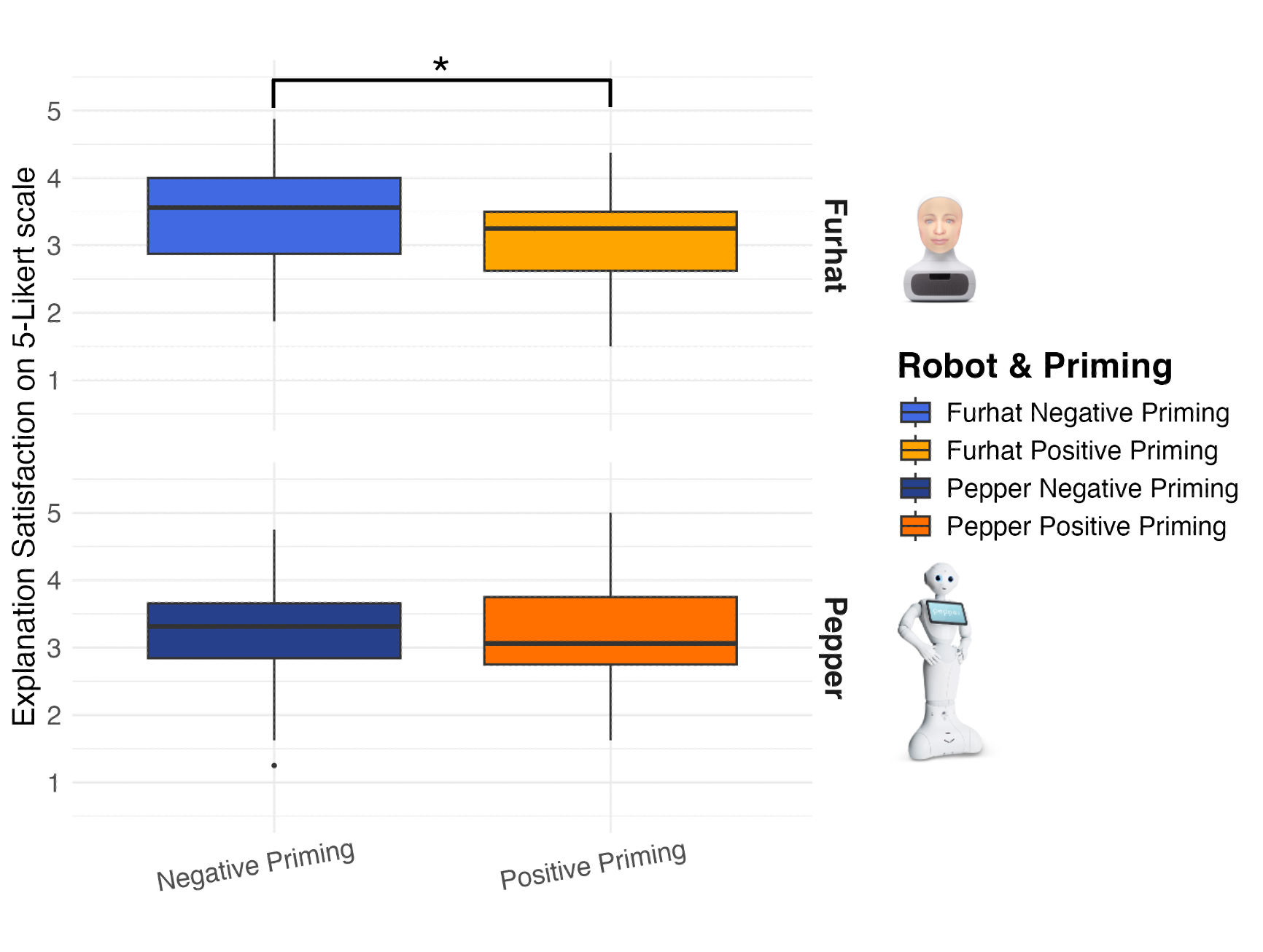}
         \caption{Explanation satisfaction results for the main study.}
         \label{fig:Explanation_main}
     \end{minipage}
     \vspace{-0.5em}
 \end{figure*}
 
\subsection{H1: Human Perception}
We conducted a Kruskal-Wallis test to compare the EmCorp subscales of \textit{expressiveness} and \textit{interpretation} across the four conditions.  
Post-hoc Mann-Whitney U tests with Bonferroni correction addressed multiple comparisons.


\textbf{Furhat:} 
A Kruskal-Wallis test revealed a statistically significant difference between the four conditions for \textit{expressiveness} ($H(3) = 15.73, p = 0.0012$), but no significant difference was found for the \textit{interpretation} ($H(3) = 4.25, p = 0.23$). 
For the ``Explanation condition'', The Mann-Whitney U test with Bonferroni adjustment between groups revealed participants perceived Furhat's \textit{expressiveness} as significantly higher ($W = 779, p = 0.00032$) when negatively primed ($Mdn = 3.0\pm1.15$) compared to positively primed ($Mdn = 3.0\pm1.15$). 
For the ``No Explanation condition'', no significant differences emerged between negative or positive priming for \textit{expressiveness} or \textit{interpretation}. 


Interestingly, negatively primed participants showed significantly higher \textit{interpretation} scores for Furhat with explanations versus no explanations, whether they were negatively primed ($ W = 1716, p = 0.04, M = 2.85, IQR = 1.67$) or positively primed ($ W = 1621.5, p = 0.03, M = 2.85, IQR = 1.28$). 
This result aligns with earlier findings that explanations help ``repair'' user impressions when expectations are low. 


\textbf{Pepper:} Under the \textit{explanation} condition, the Wilcoxon signed rank test showed no statistically significant differences in how participants perceived Pepper's \textit{expressiveness} and \textit{interpretation} whether negatively or positively primed.
The same held true under the condition of no explanation. 
Hence, for Pepper, participants' EmCorp perceptions did not appear to be modulated by either priming or explanation. One possible reason is that Pepper's more disruptive tasks (e.g., dropping objects) overshadowed any benefit derived from a brief verbal explanation. 
Figure \ref{fig:Emcorp_main} illustrates the average responses to the EmCorp subscale for both robots.

\subsection{H2: Explanation Satisfaction}
Finally, for those participants who received explanations after the robot's failure, we compared their satisfaction scores under negative vs. positive priming (see Figure \ref{fig:Explanation_main}).

\textbf{Furhat:} Participants were significantly more satisfied with the explanations ($W = 994.5, p = 0.03$) when they were negatively primed ($Mdn = 3.56\pm0.81$) than when npositively primed ($Mdn = 3.25\pm0.60$). 

\textbf{Pepper:} No significant difference was observed in explanation satisfaction between negative and positive priming. This aligns with the EmCorp findings, suggesting Pepper's physically disruptive failures may have overshadowed purely verbal explanations.

Overall, these results confirm that explanations have a stronger positive impact when initial expectations are lower, particularly for a highly expressive robot like Furhat.


%% file: sections/07_Discussion.tex
\section{Discussion}

In this paper, we evaluated the impact of robot explanations based on people's positive and negative expectations of robots. To achieve this, we primed individuals using a short video to shape their expectations about the robot's capabilities. 
Then, we evaluated how they perceived the robot's explanations regarding a mistake it made in another video they watched after being primed. 
The results confirm that the positive and the negative priming successfully led to high and low expectations from the robots, respectively. Importantly, robot explanations were helpful in improving people's perception of the robot, and explanations were even more critical when people were negatively primed.
Finally, our result showed that priming effects are real but can be robot-dependent, reflecting the malleability of user attitudes in HRI.

In terms of general attitudes toward robots, our results from the GAToRS scale—collected in both the priming study and the main evaluation indicate that these priming effects can substantially alter short-term attitudes toward robots. However, given that we measured this scale immediately after the interaction, we do not conclude whether these changes would persist over the long term. 
This online study, although controlled, provides only a snapshot of users' impressions–more ecological or longitudinal studies are necessary to see if these effects hold in real-world settings.

The robot explanations were useful to mitigate the impact of robot mistakes on user perceptions, as seen in Figure~\ref{fig:Emcorp_main} specifically for the Furhat robot (aligned with \textbf{H1}). People's evaluation of the robot's \textit{shared perception \& interpretation} were higher when the robot provided explanations (for both positive and negative priming) compared to when it did not.
However, these positive effects were not evident for Pepper, aligning with prior work that suggests embodiment can moderate how users interpret failures and explanations ~\citep{kiesler2008anthropomorphic, 10.1145/3313831.3376372}. 
This could be due to differences in the tasks performed by the two robots~\citep{relatedwork_expectation_meister2014robot}, as well as the embodiment of the Pepper robot, which may influence the effectiveness of robot explanations~\citep{kiesler2008anthropomorphic}, user expectations and how people interpret robot failures~\citep{10.1145/3313831.3376372}. 
We also note that Pepper's more disruptive failure (e.g., dropping an object) might overshadow a purely verbal explanation, making it harder for explanations alone to recover user trust.
In addition to the robots' physical embodiment, the animated nature of Pepper (e.g., full-body movement and sound) may have contributed to changes in users' perception~\citep{geva2022interaction}. It is important to highlight that although we observed significant results only for the Furhat robot during the main evaluation, the priming study demonstrated strong effects of positive and negative expectations for both robots. 
This discrepancy highlights the importance of the robot's active capabilities and the severity or visibility of its failures: minor communication lapses in Furhat versus more pronounced physical breakdowns in Pepper. 

Moving further, our results indicate that robot explanations had a stronger positive impact when participants held lower expectations of the robot's capabilities, as shown in Figure~\ref{fig:Explanation_main}. Specifically, individuals were more satisfied with the Furhat robot's explanations when they had been negatively primed (i.e., held lower expectations) compared to those who were positively primed (consistent with \textbf{H2}). This outcome is consistent with expectancy-disconfirmation theory~\citep{oliver1994outcome}, which suggests that satisfaction depends on how perceived performance compares to prior expectations.  
When expectations are initially low, an effective explanation that clarifies or justifies failures could improve users' forecast about the robot's capabilities, thus resulting in a more positive assessment. 
Furthermore, from the attribution theory perspective~\citep{weiner2010development}, explanations can shift how users assign responsibility for a robot's errors–clarifying that technical constraints, rather than incompetence, caused the failure– thus improving the perception of the robot.
These two theories offer insight into \textit{why} explanations had a higher impact on improving how the Furhat robot was perceived when participants were primed negatively–i.e., expected lower capabilities–they reduced negative disconfirmation and recalibrated users' attribution away from blaming the robot.

Aligned with these frameworks, our findings reinforce explanations as a powerful repair strategy \citep{das2021ExplainableAIRobot, lee2024rex} especially when \textit{people's expectations of robots are lower}.
When Furhat offered an explanation, participants who had been primed to expect poor performance rated the robot's expressiveness significantly higher, suggesting that clarifying the reason behind mistakes can buffer against negative judgments. These results echo prior work on how people's preconceived notions shape their interpretation of events \citep{allan2022doors, chiu1997implicit, chiu1997lay, desideri2021mind}, illustrating that user perception of robot failures is not merely a technical issue but also a social and psychological one.

Interestingly, in the case of the Pepper robot, explanations did not yield a significant improvement in satisfaction regardless of the positive or negative priming. One possibility could be linked to Pepper's additional modalities–such as whole-body movement- that might have overshadowed or minimized the effect of purely verbal explanation, especially if users expect these movements to convey cues consistent with the verbal content. 
This tension between \textit{how} explanations are delivered and \textit{how} users form mental models of the robot points to an important design consideration: matching the explanation modality to the robot's primary communication channel may enhance the explanatory impact. 
In short, our study suggests that HRI is highly contextual: robot embodiment, the nature of the task, and preconceptions formed through priming all interact to shape user attitudes. Explanations may be most effective for expressive robots or when initial user expectations are notably low.

\subsection{Limitations}

We can identify a range of limitations in our study that could be considered when interpreting the result and also when developing future research in this domain. 
\textbf{First}, our evaluation was conducted in an online setting using short video clips rather than real-world or prolonged interactions. This raises questions about ecological validity and whether the observed effects would generalize to more in-person interaction scenarios. 
\textbf{Second}, our task designs for Furhat and Pepper were not fully identical; Pepper's physically disruptive errors (e.g., dropping objects) differ qualitatively from Furhat's communication-based failures, which may have influenced how explanations were perceived. 
\textbf{Third}, we assessed user attitudes and perceptions immediately after the interactions, leaving the long-term durability of priming or explanation benefits unexplored. 
\textbf{Finally}, experiment participants were selected through an online platform who may not represent the whole society in terms of experience or interest in using robots.




\section {Future Work and Practical Takeaway}

Building on these findings, future research could pursue several directions. For instance, longitudinal or real-world studies could investigate whether short-term effects or explanations of benefits persist once users gain more hands-on experience with robots. Studying additional robot embodiments under standardized tasks may further clarify how morphology and interaction style impact perceived failures and explanations. Examining multimodal explanation strategies that combine speech with synchronized gestures or on-screen text might help mitigate disruptive robot errors, especially for more mobile robots.

Based on our findings, we developed the following practical takeaways that could help HRI researchers develop future research in this domain. 
\begin{itemize}
    \item \textbf{Context Matters:} Our findings underscore the importance of managing user expectations in different contexts (e.g., education, healthcare, customer service). To understand the role of context better, designing consistent failure scenarios and explanation strategies is crucial.
    \item \textbf{Explanations Are Most Effective for Low Expectations:} Participants with negative priming (low expectations) saw the largest gain from an explanation. This suggests that HRI designers should pay special attention to ``damage control'' in contexts where users might doubt the robot's competence from the start (e.g., healthcare).
    \item \textbf{Embodiment and Modality Must Align:} For Furhat, verbal explanations paired well with its expressive facial features. Pepper's more complex movements and physically disruptive errors might need more integrated or multimodal explanations to rebuild trust effectively.
    \item \textbf{Task Complexity and Failure Severity:} Pepper's tasks involved larger, more visible mistakes (e.g., dropping objects), which might overshadow the explanations. Systematic comparisons of task types and failure severity can help generalise these findings.
\end{itemize}

\section {Conclusion}

In this paper, we presented a two-stage investigation including a \textbf{priming evaluation} that validated how brief interaction videos of a robot displaying its capable side (positive priming) versus failure-prone side (negative priming) could reliably induce high or low expectations in participants. We investigated this effect for two distinct robot embodiments (Pepper and Furhat). After verifying whether the videos were successful in setting the expectations as we planned, we conducted a \textbf{main evaluation} showing how \textit{explanations} significantly influence user perception, especially when initial expectations are low. The different observed outcomes for Pepper versus Furhat highlight the interplay of embodiment, task design (in terms of complexity), failures (in terms of severity), and communication modalities in shaping the efficacy of robot explanations. Overall, our findings emphasize the importance of managing user expectations and employing tailored explanation strategies to sustain trust and satisfaction in human-root interaction–even (or especially) when mistakes inevitably occur.

%% file: sections/09_Acknowledgment.tex
\section*{Acknowledgment}
This work was partially funded by the S-FACTOR project from NordForsk, the Vinnova Competence Center for Trustworthy Edge Computing Systems and Applications at KTH, and the Wallenberg Al, Autonomous Systems and Software Program (WASP) funded by the Knut and Alice Wallenberg Foundation.  Beatriz
Nogueirab and Tiago Guerreiro were partially supported by the LASIGE Research Unit, ref. UID/00408/2025. Financial support for Shelly Levy Tzedek was provided by the Rosetrees Trust and by the Consolidated Anti-Aging Foundation.